\documentclass{article}
\usepackage{arxiv}

\usepackage{times}
\usepackage{soul}
\usepackage{url}
\usepackage[hidelinks]{hyperref}
\usepackage[utf8]{inputenc}
\usepackage[small]{caption}
\usepackage{graphicx}
\usepackage{amsmath}
\usepackage{amsthm}
\usepackage{booktabs}
\usepackage{algorithm}
\usepackage{algorithmic}
\usepackage[switch]{lineno}
\usepackage{amssymb}
\usepackage{multirow}

\title{Graph based Environment Representation for Vision-and-Language Navigation in Continuous Environments}

\author{
Ting Wang$^{1,2}$
\and
Zongkai Wu$^{2,*}$\and
Feiyu Yao\And
Donglin Wang$^{2,*}$
\affiliations
$^1$Zhejiang University\\
$^2$Westlake University
\emails
\{wangting, wuzongkai\}@westlake.edu.cn,
feiyu.yao@columbia.edu,
wangdonglin@westlake.edu.cn
}

\begin{document}

\maketitle

\begin{abstract}
    Vision-and-Language Navigation in Continuous Environments (VLN-CE) is a navigation task that requires an agent to follow a language instruction in a realistic environment. The understanding of environments is a crucial part of the VLN-CE task, but existing methods are relatively simple and direct in understanding the environment, without delving into the relationship between language instructions and visual environments. Therefore, we propose a new environment representation in order to solve the above problems. First, we propose an Environment Representation Graph (ERG) through object detection to express the environment in semantic level. This operation enhances the relationship between language and environment. Then, the relational representations of object-object, object-agent in ERG are learned through GCN, so as to obtain a continuous expression about ERG. Sequentially, we combine the ERG expression with object label embeddings to obtain the environment representation. Finally, a new cross-modal attention navigation framework is proposed, incorporating our environment representation and a special loss function dedicated to training ERG. Experimental result shows that our method achieves satisfactory performance in terms of success rate on VLN-CE tasks. Further analysis explains that our method attains better cross-modal matching and strong generalization ability. 
    
\end{abstract}

\section{Introduction}
    Vision-and-language navigation (VLN) \cite{anderson2018vision} is a task that requires an agent to navigate from an arbitrary initial position to a described destination by understanding real-time image information and language instructions in a photo-realistic 3D environment. 
    In recent years, many studies \cite{wang2018look,fried2018speaker,wang2019reinforced,ma2019selfmonitoring} have done model improvement and method innovation on this basis to achieve satisfactory results.
    However, some perfect assumptions of VLN are far from real robots, including hopping navigation, known environment topology, and precise localization.
    Therefore, the Vision-and-Language Navigation in Continuous Environments (VLN-CE)~\cite{krantz2020beyond} task is proposed to provide a more realistic platform for VLN robot by performing a set of low-level actions. 
    This means that the agent in continuous environments needs to have better environment understanding capability. 
    
    \begin{figure}[t]
    	\includegraphics[scale=0.45]{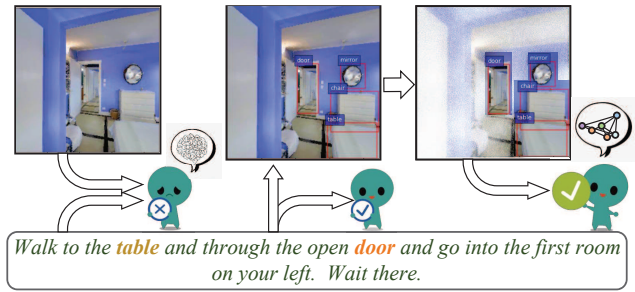}
    	\centering
    	\caption{For VLN-CE, the agent needs to complete an instruction in a real environment. Without sufficient understanding of the environment, the agent can easily become confused about the instruction. When the agent has a good understanding of the environment and is aware of the objects and their relationships, the navigation task is easy to achieve.} 
    	\label{fig1}
    \end{figure}
    
    At present, the work on the VLN-CE task is still relatively lacking, and there is a large space for improvement in performance. 
    The VLN-CE dataset~\cite{krantz2020beyond} is obtained by converting the navigation trajectories of the Room-to-Room dataset~\cite{anderson2018vision} into the Habitat Simulator~\cite{savva2019habitat}. 
    Visual information in the dataset is real image captured by the camera.
    \cite{irshad2021sasra} propose a cross-modal Semantic-Linguistic Attention Map Transformer.
    \cite{krantz2021waypoint}, \cite{raychaudhuri2021language} and \cite{hong2022bridging} convert low-level action predictions into waypoint predictions in different ways, achieving the best current navigation performance.
    But they do not realize the importance of understanding the environment and the intrinsic connection between the visual environment and language.
    These problems make it impossible for the agent to achieve a high-level matching of vision and language, thus limiting the navigation performance. 
    
    For example, as shown in Figure \ref{fig1}, the language instruction “Walk to the table and through the open door and go into the first room on your left. Wait there.” requires the agent to understand the language and interact with the environment to complete the navigation task. 
    Most of previous works directly extract the visual feature from image to make navigation decision. 
    However, referring to humans, the objects such as “table", “door", and “room" should be first identified in sequence, and then the navigation decision is made by understanding the relationship between object-object and object-human.
    Such operations require the agent to be able to detect the objects in vision and establish an environment representation in semantic level.
    Finally, the agent should make navigation decision by matching this representation and language instructions.
    
    In order to achieve the above idea, we propose a new environment representation.
    First, we analyze all object descriptions in language instructions and summarize a vocabulary.
    Then, we introduce object detection technology, determine the target to be detected according to the summarized vocabulary, and identify the objects in the environment within the agent's field of vision.
    To express the environment, we propose an Environment Representation Graph (ERG) with the detection results. Each node information of ERG includes direction information between object and agent, one-hot label vector and confidence.
    Such operation can improve the agent's environment understanding capacity under language conditions, and can relatively reduce the gap caused by the texture difference among different buildings.
    
    However, to express the environment, it is not enough to simply introduce semantic information. It is also necessary to know the relationship of object-object and object-agent, that is, the expression of edges in ERG.
    Whether manually defining discrete relationship symbols or estimating distances, there are problems such as high labor costs or missing information.
    Therefore, we consider Graph Convolution Network (GCN) \cite{satorras2018few} to learn the relationship between nodes.
    That is, our ERG neither requires a pre-defined adjacency matrix nor relies on additional external knowledge, but learns relational representations from the training dataset with the help of object detection knowledge.
    Sequentially, we obtain a continuous expression about ERG, which can imply more information and can be flexibly adjusted according to the needs of the agent.
    Finally, we multiply the ERG expression as the attention map by the object label embedding to obtain the final environment representation at the semantic level.
    It is worth noting that in order to further alleviate the expression differences between navigation instructions and object detection labels, we use TinyBert to transform them into the same embedding space.
    
    After that, we design a cross-modal attention navigation framework. Our framework gets three inputs about the environment: our environment representation, RGB and depth, and utilizes the GRU to carry the overall information flow. The historical context encoded by the GRU, in addition to the previous RGB-D, will also include the environment representation processed by soft-attention. In the action selection module, our environment representation features are also added to make decisions. Our cross-modal attention framework not only improves the connections between multi-modal information, but also amplifies the influence of our environment representations on navigation decisions.
    In the training phase, in addition to considering the navigation cross-entropy loss, we also propose a new loss function that considering the spatial consistency of the relationship between the same pair of objects when the perspective changes. 
    Overall, the main contributions of this article are summarized as follows:
    \begin{itemize}
    	\item We propose an environment representation for VLN-CE. We introduce semantic information and construct the ERG in real time through object detection, where the relational representation is learned by GCN. 
    	
    	\item We propose a new cross-modal attention VLN-CE framework combined with an environment representation module. In the training phase, We consider a combination of the two losses, the navigation cross-entropy loss and the spatial consistency loss of the relationship between the same pair of objects when the perspective changes. 
    	
    	\item The experimental results demonstrate that our graph based environment representation indeed helps to improve the navigation performance. Better cross-modal matching is attained and the generalization of the navigation model is enhanced in unseen environments. 
    \end{itemize}
    
    \begin{figure*}[t]
    	\includegraphics[scale=0.61]{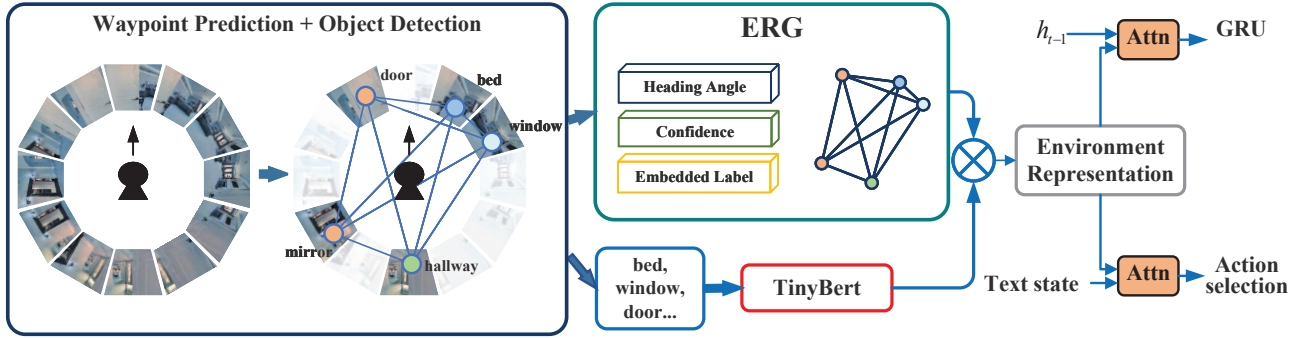}
    	\centering
    	\caption{Illustration of environment representation. The waypoint predictor predicts k candidate directions for the agent. Then ERG is generated by extracting object one-hot label features, confidence and orientation through object detection from $K$ candidate waypoint views as node information. To further facilitate the matching of environment and language, ORG is used as the attention map to encode the TinyBERT embedding of the label to obtain the final environment expression.} 
    	\label{fig2}
    \end{figure*}

\section{Related Work}
\paragraph{Vision-and-Language navigation}
    The Vision-and-Language Navigation (VLN) task proposed by \cite{anderson2018vision} has attracted extensive research attention in recent years. Many works make improvements and innovations on this basis. 
    \cite{wang2018look} first combine model-free and model-based deep reinforcement learning for VLN. 
    Speaker-Follower \cite{fried2018speaker} performs data augmentation by generating language instructions for efficient navigation. 
    To address the cross-modal grounding and information feedback issues, \cite{wang2019reinforced} propose a novel Reinforced Cross-Modal Matching (RCM) approach. 
    Similarly, the Regretful Agent \cite{ma2019regretful} and Tactical Rewind backtrack \cite{ke2019tactical} introduce a self-monitoring mechanism and global signals, respectively. 
    Referring to previous methods
    , \cite{zhu2020vision} introduce four self-supervised auxiliary reasoning tasks. 
    \cite{wu2021improved} propose a new transformer-based multimodal framework for navigator and speaker, respectively.
    \cite{zhang2021curriculum} propose a VLN model based on curriculum learning.
    ENVEDIT \cite{li2022envedit} and REM \cite{liu2021vision} perform data augmentation by editing environment.
    In addition, there are many works \cite{liang2022visual,majumdar2020improving,hong2021vln,qi2021road,hao2020towards,guhur2021airbert} to improve VLN performance through various pre-training methods.
    \cite{jain2019stay} propose a new dataset Room-for-Room (R4R) and a new metric Coverage weighted by Length Score (CLS). 
    \cite{ku2020room} introduce a multilingual Room-Across-Room (RxR) dataset.
    \cite{vln-pano2real} transfer the VLN task from the simulation environment to the physical robotic platform.
    \cite{krantz2020beyond} eliminate unrealistic assumptions about sparse navigation graphs and propose the VLN-CE task.
    \cite{krantz2021waypoint,raychaudhuri2021language,hong2022bridging} propose waypoint-based VLN-CE navigation model in different ways. 
    \cite{wang2021visual} introduce a meta-learning-based visual perception generalization strategy that enables agents to adapt to new sensor configurations.
    \cite{irshad2021sasra} propose a cross-modal Semantic-Linguistic Attention Map Transformer.

\paragraph{Graph for navigation}
    In the field of intelligent navigation, there are some works that have been combined with graph and achieved good results. 
    There are the following researches in the field of object navigation.
    \cite{du2020learning} propose three complementary techniques: an object relational graph, a trial-driven imitation learning and a memory-augmented tentative policy network.
    Similarly, \cite{xiaobo2021agent} introduce an Agent-Centric Relation Graph (ACRG) for learning visual representations.
    \cite{gadre2022continuous} propose scene representations suitable for different downstream tasks.
    An DOA graph \cite{dang2022unbiased} explicitly learns attention relationships between objects and allocates more reasonable attention resources to object features and global image features.
    A novel two-layer hierarchical reinforcement learning method with a Goals Relational Graph \cite{ye2021hierarchical} is proposed.
    \cite{zhang2021hierarchical} propose an online learning mechanism based on a hierarchical object-to-zone (HOZ) graph.
    As for VLN-related work, a Structured Scene Memory (SSM) \cite{wang2021structured} is proposed to capture visual and geometric cues in the environment, and is equipped with a collect-read controller for information gathering and long-term navigation reasoning. 
    \cite{hong2020language} propose a Language and Visual Entity Relationship Graph and a message-passing algorithm.
    \cite{zhu2021soon} introduce a Scenario Oriented Object Navigation (SOON) task and propose a graph-based exploration method, which models navigation states as graphs and stabilizes training by learning sub-optimal trajectories.
    Unlike them, we will realize the environment representation at the semantic level through building a graph.

\section{Methodology}
    In this paper, we aim to create our environment representation for introducing rich semantic and contextual information, which enables understanding and reasoning of environments and facilitates cross-modal matching. 
    We start by describing the VLN-CE task and the specific learning process of the ERG-centered environment representation.
    Then we show overall VLN-CE framework incorporating our environment representation.
    Finally, we introduce the training and loss of the whole model.

\subsection{Task Definition}
    At each time step $ t $, the VLN-CE agent receives the instruction $ L $, the RGB observation $ \mathcal{V}^{rgb}_t $ and the depth observation $ \mathcal{V}^{depth}_t $.
    The language instruction $ L $ is a sentence composed of $ M $ words.  
    Visual observations $ \mathcal{V}^{rgb}_t $ and $ \mathcal{V}^{depth}_t $ received in real time are egocentric RGBD images from the simulator with a resolution of $ 256 \times 256 $ and a horizontal field-of-view of $ 90 $ degrees. 
    Then the agent predicts the next action $ a_t $ based on the current state and the above information (vision and language). 
    In this paper, we add our ERG-centric environment representation for decision making.
    The entire navigation can be viewed as a Partially Observable Markov Decision Process (POMDP).
    The action space $ \mathcal{A} $ of all agents consists of four simple, low-level actions - Move forward $ 0.25 $m, Turn left or Turn right $ 15 $ degrees, and Stop. 
    The agent repeats this process and performs a series of actions $ a =\{a_0, a_1, \dots, a_t\}$ until the “Stop" action is selected.
    The success of the VLN-CE task can be declared only if the agent's position is close enough to the instruction target when it makes the “Stop" action.
    
\subsection{Environment Representation}    
    In this section, we explain how an ERG can be constructed in real-time to obtain richer environment representations for navigation.
    We first discuss building local node representations by object detection. 
    Then we discuss how to obtain relational representations in ERG with GCN. 
    Finally, we take the output of the ERG as an attention map, and combine it with the label embedding to get the final environment representation. 
    A general illustration of the environment representation is shown in Figure \ref{fig2}.
    
    \textbf{Detection for Node Representation}
    
    In order to obtain a more contextual environment representation, we first need to detect all objects from the environment. 
    We achieve this by using the Faster RCNN \cite{ren2015faster} pretrained on Visual Genome \cite{krishna2017visual} as our object detector. 
    As shown in Figure \ref{fig2}, at each time step, we localize all objects of interest through Faster RCNN for the egocentric RGB images of $ k $ candidate waypoints. 
    In this paper, we analyze all object descriptions in language instructions and summarize a vocabulary, and pre-set $ U = 100 $ object categories of interest based on instructions. 
    Then we define a graph $ \mathcal{G} = (\mathcal{N},\mathcal{E}) $ to construct our ERG.
    Each node $ n \in \mathcal{N} $ represents an entity category in the environment, and each edge $ \varepsilon \in \mathcal{E} $ represents a relationship between two entities.
    According to each detection result, the heading feature $ d_{j} $, the confidence $ q $ and a one-hot encoded label vector $ r $ will serve as a local node representation of our ERG. 
    $ d_{j} $ is a heading feature vector consisting of $ [ \mbox{sin}\phi; \mbox{cos}\phi ] $, where $ \phi $ is the heading angle of the candidate waypoint relative to the agent.
    Each time, we aggregate all detections in $ k $ directions and construct only one ERG, as described in Algorithm \ref{al1}, lines 4 to 11.
    These operations construct the ERG to make it easier for an agent to understand the environment, and introduce semantic information that can alleviate the domain gap caused by style and texture differences among different buildings to a large extent.
    
    \begin{figure}[b]
    	\includegraphics[scale=0.35]{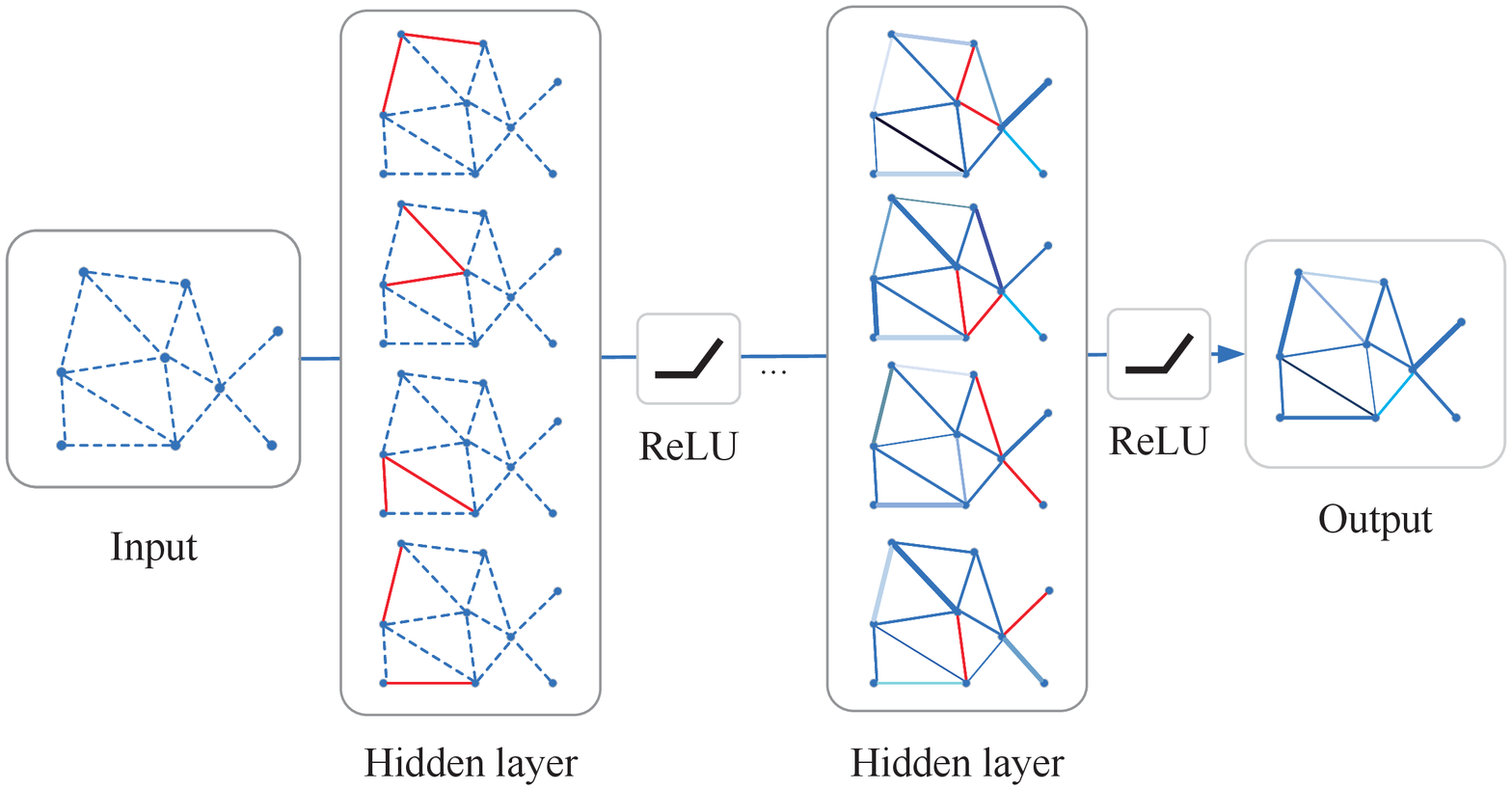}
    	\centering
    	\caption{The learning process of ERG. } 
    	\label{fig3}
    \end{figure}
    
    \textbf{Learning for Our ERG}
    
    Regarding the expression of edges, we have considered many possibilities. 
    For example, artificially pre-defined discrete relationship symbols, such as “besides", “above", “inside", etc. 
    On the one hand, this method is time-consuming and labor-intensive, and on the other hand, it may lose a lot of important information such as the specific performance of relative positions or metric relationships, etc. 
    Alternatively, using a distance metric as a relational expression also produces errors while losing information, because our perspective changes in real time and our navigation instructions are not guided by distance. 
    Therefore, we consider applying GCN to acquire continuous representations of edges in a learned manner.
    
    Our GCN takes all nodes $ X \in \mathbb{R}^{| U |\times D} $ as input, and then embeds each input node with a matrix $ W_{g} \in \mathbb{R}^{D \times U} $, where $ D $ is the dimension of the node features. Afterwards, our GCN embeds all nodes through the adjacency matrix $ E \in \mathbb{R}^{| U |\times U} $, and outputs a new encoding $ \mathcal{Z} \in \mathbb{R}^{| U |\times U} $, which expresses the environment, including the objects in the environment, the relationship between objects, and the relationship between objects and the agent.
    Executing line 12 of Algorithm \ref{al1}, our proposed ERG is expressed as: 
    \begin{equation}\label{eq1}
    	\mathcal{Z} = f(E \cdot X \cdot W_{g})
    \end{equation}
    where $ f(\cdot) $ represents the ReLU activation function. Note that our model learns both the node embedding $ W_{g} $ and the adjacency matrix $ E $. 
    This flexible representation of edges contains rich spatial relationships between category entities. 
    The learning process of ERG is shown in Figure 3. 
    Because our ERG is learned, it can be adjusted during navigation according to the actual needs of the agent and can also be adapted to different environments and tasks. 

    \begin{algorithm}[t]
    	\caption{Establish environment representation.}
    	\label{al1}
    	\begin{algorithmic}[1]
    				
    		\FOR {Each $ t $-th epochs}
    		\STATE Obtain $ k $ candidate RGB images through waypoint predictor.
    		\STATE Build ERG  $\mathcal{G}$ with empty node.
    		\FOR {$j=1:k$}
    		
    		\STATE Detect objects by Faster RCNN from $j$-th candidate image.
    		
    		\FOR {$ j=1:k $}
    		\IF {$q$ $>$ $\mathcal{G}$[object][confidence]}
    		\STATE Update $\mathcal{G}$[object] with the direction information between object and agent, one-hot label vector and confidence.
    		\ENDIF
    		\ENDFOR
    		\ENDFOR
    		\STATE Learn the representation $\mathcal{Z}_t$ of ERG using GCN by equation \ref{eq1}.
    		\STATE Using $\mathcal{Z}_t$ as attention map to the label embedding expression $\mathcal{D}$.
    		\STATE Output environment representation $\mathcal{O}_t$  as equation \ref{eq2}.
    		
    		\ENDFOR
    	\end{algorithmic}	
    \end{algorithm}
    
    \textbf{Attention for Environment Representation}
    
    In order to make the agent more focus on the object of interest and further enhance the matching ability of environment and instructions, we use the attention mechanism to generate our final environment representation. 
    First, TinyBERT is used to unify the language embedding of detection labels and navigation instructions into the same space, and obtain the label embedding expression $ \mathcal{D} \in \mathbb{R}^{|U| \times m} $. Then we use $ \mathcal{Z} $ as the attention map to label embedding. Corresponding to the last two steps in Algorithm \ref{al1}, our graph attention is expressed as:
    \begin{equation}\label{eq2}
    	\mathcal{O}_{t} = f(\mathcal{Z}_{t} \cdot \mathcal{D}) 
    \end{equation}
    and there are no learnable parameters here. $ \mathcal{O}_{t} $ is the semantic-level environment representation. 

\subsection{Overview VLN-CE Framework}
\begin{figure}[t]
    	\includegraphics[scale=0.42]{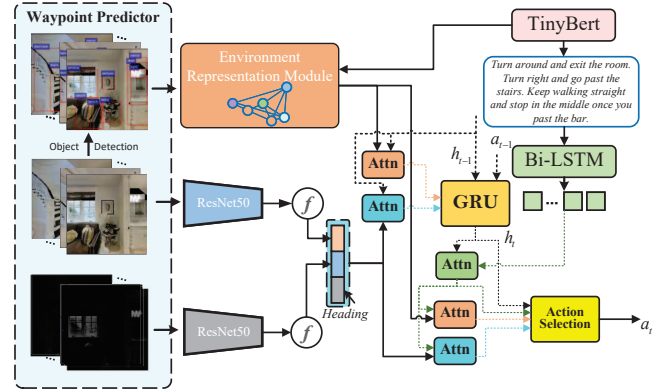}
    	\centering
    	\caption{Overview of our cross-modal attention VLN-CE framework. Our framework consists of a waypoint predictor, an environment representation, a visual representation, a language understanding and an action decision module.} 
    	\label{fig4}
    \end{figure}

    Figure \ref{fig4} depicts the framework of our complete cross-modal attention navigator.
    In this paper, we refer to \cite{hong2022bridging}, which predicts candidate waypoints in advance to narrow the action space and bridge the gap between discrete and continuous environments.
    First, the candidate waypoint predictor uses only RGB-D visual information to estimate $ k $ navigable positions for the agent in a continuous environment at each time step. The process is similar to construct a local navigable graph centered on the agent. 
    
    \textbf{For vision}, we use two ResNet50s \cite{he2016deep} to encode RGB and depth observations separately, one pre-trained on ImageNet \cite{russakovsky2015imagenet} for classification and one pre-trained on Gibson \cite{xia2018gibson} for point-goal navigation. 
    We denote the encoded visual observations as $ \{\mathcal{V}^{rgb}_{t,j}\}_{j=1}^k \in \mathbb{R}^{2048} $ for RGB and $ \{\mathcal{V}^{depth}_{t,j}\}_{j=1}^k \in \mathbb{R}^{128} $ for depth. 
    Then, at each time step, we fuse the RGB, depth and heading corresponding to each candidate waypoint to get the final visual information as
    \begin{equation}\label{1}	
    	\mathcal{V}_{j} = [\mathcal{V}^{rgb}_{j} W_{rgb}; \mathcal{V}^{depth}_{j} W_{depth}; d_{j}] W_{fuse}
    \end{equation}
    where $ W $ are learnable linear projections and $ d_{j} $ is a vector encoding representing the relative orientation of the $ j $th candidate viewpoint. 
    
    \textbf{For environment representation}, previous works only consider RGB and depth, which cannot allow the agent to understand the environment in detail and makes it difficult to achieve a high degree of matching between vision and language. So we propose a richer environment representation based on ERG. 
    At each time step, we construct the ERG $ \mathcal{G}_t $ and output the graph encoding $ \mathcal{Z}_t $ using GCN. 
    Then we use $ \{\mathcal{Z}_{t,u}\}^{U}_{u=1} $ as the attention map to the detection label to obtain the final environment representation $ \{\mathcal{O}_{t,u}\}^{U}_{u=1} $, where $ U $ is the number of nodes.
    
    \textbf{For language}, we use Tinybert to get embeddings $ L=\{l_1, l_{2}, \dots, l_M\} $ for each instruction of length $ M $. A bidirectional LSTM \cite{hochreiter1997long} is used to encode $ L $ as
    \begin{equation}\label{}
    	\mathcal{C} = \{c_{1}, c_{2}, \dots, c_{M}\} = \mbox{Bi-SLTM} (l_1, l_{2}, \dots, l_M)
    \end{equation}
    
    \textbf{For navigation}, we consider CMA \cite{wang2019reinforced} as our policy network. 
    As the navigation progresses, the agent's visual perception changes accordingly. Attention-based historical trajectory encoder GRU encodes the historical state of the agent as
    \begin{equation}\label{}
    	h_{t} = \mbox{GRU} ([\bar{\mathcal{O}_{t}},\bar{\mathcal{V}_{t}}, a_{t-1} W_{action}], h_{t-1})
    \end{equation}
    where $ W_{action} $ is a learnable linear projection, $ a_{t-1} $ is the action taken at the last time step, $ h_{t-1} $ is the historical state context of the last time step, $ \bar{\mathcal{V}_{t}} = \sum_{j=1}^{k} \alpha_{t, j} \mathcal{V}_{t,j} $ is the weighted sum of the visual features after attention, and $ \bar{\mathcal{O}_{t}} = \sum_{u=1}^{U} \beta_{t, u} \mathcal{O}_{t,u} $ is the weighted sum of environment representation after attention,
    Specifically, we utilize the dot product attention to obtain $ \bar{\mathcal{V}_{t}} $ and $ \bar{\mathcal{O}_{t}} $,
    \begin{equation}
    	\begin{aligned}
    		\bar{\mathcal{V}_{t}} &= \mbox{Attn}(h_{t-1},\{ \mathcal{V}_{t,j} \}_{j=1}^{k})\\
    		&= \sum \limits_{j} \mbox{softmax} (h_{t-1}W^{(1)}_{h}(\mathcal{V}_{t,j} W_{\mathcal{V}})^{\mbox{T}})\mathcal{V}_{t,j}.
    	\end{aligned}
    \end{equation}
    \begin{equation}
    	\begin{aligned}
    		\bar{\mathcal{O}_{t}} &= \mbox{Attn}(h_{t-1},\{ \mathcal{O}_{t,u} \}_{u=1}^{U}) \\
    		&= \sum \limits_{u} \mbox{softmax} (h_{t-1}W^{(2)}_{h}(\mathcal{O}_{t,u} W_{\mathcal{O}})^{\mbox{T}})\mathcal{O}_{t,u}.
    	\end{aligned}
    \end{equation}
    where $ W $ are learnable linear projections.
    
    We condition on the visual historical state to locate the language instruction of current interest. The vision-conditioned textual feature is computed at each time step based on the historical state $ h_{t} $ as
    \begin{equation}\label{}
    	\widehat{\mathcal{C}_{t}} = \mbox{Attn} (h_{t}, \{c_{i}\}_{i=1}^{M}) .
    \end{equation}
    
    Then we look for visual and semantic emphasis based on language. The text-conditioned visual feature is computed at each time step based on the textual feature $ \widehat{\mathcal{C}_{t}} $ as
    \begin{equation}\label{}
    	\widehat{\mathcal{V}_{t}} = \mbox{Attn} (\widehat{\mathcal{C}_{t}}, \{\mathcal{V}_{j}\}_{j=1}^{k}) .
    \end{equation}
    
    Similarly, The text-conditioned environment feature is computed as
    \begin{equation}\label{}
    	\widehat{\mathcal{O}_{t}} = \mbox{Attn} (\widehat{\mathcal{C}_{t}}, \{\mathcal{O}_{u}\}_{u=1}^{U}) .
    \end{equation}
    
    Finally, we perform action selection based on historical state $ h_{t} $, textual feature $ \widehat{\mathcal{C}_{t}} $, visual feature $ \widehat{\mathcal{V}_{t}} $ and environment feature $ \widehat{\mathcal{O}_{t}} $ to decide which candidate waypoint to choose. 
    The probability of each waypoint is calculated as 
    \begin{equation}\label{}
    	p_{t,j} = \mbox{softmax} ([h_{t}, \widehat{\mathcal{C}_{t}},	\widehat{\mathcal{V}_{t}}, \widehat{\mathcal{O}_{t}} ] W_{a} (\mathcal{V}_{t,j},W_{v})^{T}) .
    \end{equation}
    The agent chooses the candidate direction with the highest probability and navigates to that location.

\subsection{Training and Loss}    
    Since the environment is static, object-to-object relationships should not change as the agent's perspective changes. 
    We hope that, at adjacent moments, the learned representation of relationship between the same pair of objects is as consistent as possible.
    We enhance the similarity of relational expressions by optimizing a consistency loss:
    \begin{equation}\label{}
    	\mathcal{L}_{G} = -\mbox{log}\frac{\mbox{exp}(\mbox{CosSim}(e,e^{+})/\tau)}{\sum \mbox{exp}(\mbox{CosSim}(e,e^{*})/\tau)} 
    \end{equation}
    where $ e $ is the feature representation of the edge in the ERG at the current moment, $ e^{+} $ is the representation of the same relationship at adjacent moments, each $ e^{*} $ is the relationship expression between all objects, $ \tau $ is the scaling parameter, and $ \mbox{CosSim}(\cdot) $ is L2 Dot product between normalized features. The closer the relationship is expressed, the higher the CosSim value.
    By optimizing this objective, we encourage the relationship expression of the same pair of objects in different perspectives to be consistent and different from other relationships.
    
    The agent's navigation ability is trained by imitation learning with a cross-entropy loss as
    \begin{equation}\label{}
    	\mathcal{L}_{IL} = -\sum \limits_{t} a_{t}^{*} \mbox{log} (p_{t})
    \end{equation}
    where $ p_{t} $ is the action probability and $ a_{t}^{*} $ is the oracle action.
    
    The overall loss function of our model can be written as
    \begin{equation}\label{}
    	\mathcal{L} = \mathcal{L}_{IL} + \mathcal{L}_{G} .
    \end{equation}

\section{Experiments}    
\subsection{Experimental Setup}
\subsubsection{VLN-CE Dataset} 
    VLN-CE dataset \cite{krantz2020beyond} is employed to evaluate our cross-modal attention with graph (Graph-CMA) framework. It contains 4475 trajectories. Each trajectory provides ego-centered images from Habitat simulator \cite{savva2019habitat}, three natural language instructions and a pre-computed shortest path via low-level actions. The dataset is split into train, validation-seen and validation-unseen sets. Validation-seen dataset shares the same scenes with training set. Environments in validation-unseen dataset are not exposed to the agent. 

\subsubsection{Metrics}
    Six metrics are established to verify the effectiveness of our VLN framework: 
    
    • \textbf{TL} (Trajectory Length) measures the average length of the predicted trajectories in navigation. 
    
    • \textbf{NE} (Navigation Error) measures the average distance (in meter) between the agent’s stopping position in the predicted trajectory and the goal in the reference trajectory. 
    
    • \textbf{nDTW} (normalized Dynamic Time Wraping) measures the normalized cumulative distance between reference path and agent position. 
    
    • \textbf{OSR} (Oracle Success Rate) is the proportion of the closest point in the predicted trajectory to the target in the reference trajectory within a threshold distance. 
    
    • \textbf{SR} (Success Rate) is the proportion of the agent stopping in the predicted route within a threshold distance of the goal in the reference route. 
    
    • \textbf{SPL} (Success weighted by inverse Path Length) is a comprehensive metric method integrating SR and TL that takes both effectiveness and efficiency into account. 
    
    In our discussion, we will primarily verify the performance of our method through SR and SPL.

\subsubsection{Implementation Details}
    We implement our agent on the Habitat simulator. After each step, the agent will obtain the images of surroundings. The semantic information is generated referring to Faster RCNN pretrained with Visual Genome + Res101 and PyTorch \cite{ren2015faster}. The instructions and semantic information are embedded by pretrained TinyBert \cite{jiao2020tinybert}. The visual information is extracted from RGB and Depth images through Resnet50 (pre-trained on ImageNet \cite{russakovsky2015imagenet} and Gibson \cite{xia2018gibson} separately). The waypoint predictor will predict at most 6 candidate waypoint views in each step. Then an ERG is established from the candidate waypoint views' semantic information. The GCN for updating ERG is composed of 5 hidden layers.
    
    \begin{table*}[t]\small
    	\centering
    	\begin{tabular}{c|cccccc|cccccc}
    		\hline
    		\multirow{3}{*}{Methods} & \multicolumn{6}{c|}{Val-seen}       & \multicolumn{6}{c}{Val-unseen}        \\
    		& TL$\downarrow$   & NE$\downarrow$   & nDTW$\uparrow$ & OSR$\uparrow$ & SR$\uparrow$ & SPL$\uparrow$ & TL$\downarrow$   & NE$\downarrow$   & nDTW$\uparrow$ & OSR$\uparrow$ & SR$\uparrow$ & SPL$\uparrow$  \\ \hline
    		CMA                      & 9.26 & 7.12 & 54   & 46  & 37 & 35  & 8.64 & 7.37 & 51   & 40  & 32 & 30 \\
    		LAW                      & -    & -    & 58   & -   & 40 & 37  & -    & -    & 54   & -   & 35 & 31 \\
    		SASRA                    & 8.89 & 7.17 & 53   & -   & 36 & 34  & 7.89 & 8.32 & 47   & -   & 24 & 22 \\
    		Waypoint models          & \textbf{8.54} & 5.48 & -    & 53  & \textbf{46} & \textbf{43}  & \textbf{7.62} & 6.31 & -    & 40  & 36 & 34 \\ \hline
    		*BG-CMA                  & 13.9 & 5.16 & 56   & 59  & 43 & 38  & 9.21 & 6.55 & 55   & 42  & 36 & 32 \\
    		*Our (Graph-CMA)         & 11.8 & \textbf{5.04} & \textbf{59}   & \textbf{61}  & \textbf{46} & 42  & 9.96 & \textbf{6.20} & \textbf{56}   & \textbf{48}  & \textbf{39} & \textbf{35} \\ \hline
    	\end{tabular}
    	\caption{Experimental results showing the performance of baselines and our method. (* represents that the results are trained in our computation environment).}
    	\label{Experiment1}
    \end{table*} 

\subsubsection{State-of-Art Baselines}    
    To elaborate the effectiveness of our proposed environment representation and our new cross-modal attention navigation framework, we compare our method with other state-of-art VLN-CE methods. Specifically, methods are as followed.
    
    \textbf{CMA} \cite{wang2019reinforced}:This method proposes a sequence to sequence model which enables the agent to have cross-modal attention and spatial visual reasoning ability. 
    
    \textbf{SASRA} \cite{irshad2021sasra}: This method designs for the agent a hybrid transformer recurrent cross-modal model focusing on aligning top-down local ego-centric semantic mapping with language.
    
    \textbf{LAW}\cite{raychaudhuri2021language}: This method guides the agent to learn policy with a language-aligned supervision scheme and a metric which measures the sub-instructions the agent has completed during navigation. 
    
    \textbf{Waypoint Model} \cite{krantz2021waypoint}: This method provides a language-conditioned waypoint prediction network for the agent. Then the agent picks waypoint referring to its hidden states and environmental features.
    
    \textbf{BG-CMA} \cite{hong2022bridging}: This method discretizes the continuous environment and provide candidate waypoints for the agent. Agent focuses on choosing candidate direction referring to a reinforced Cross-Modal Matching approach. 
    
\subsection{Main Results}    
    We validate the agent with our proposed Graph-CMA on VLN-CE dataset. As can be seen in Val-seen, our method achieves best performance in almost all metrics. The results are in Table \ref{Experiment1}, which is divided into two parts. The upper part shows the claimed performance of several more classic methods in the field of VLN-CE. 
    In the comparison among results, Waypoint models outperforms other three methods in almost all metrics.
    In val-unseen dataset, the improvements of our Graph-CMA in SR and SPL are 3$\%$ absolute (8.3$\%$ relative) and 1$\%$ absolute (3$\%$ relative) compared with Waypoint models.
    In lower part of the table, we focus on comparing the latest research (BG-CMA) results with our method. 
    BG-CMA \cite{hong2022bridging} is claimed to outperform Waypoint models in all metrics except TL metric.
    It is worth noting that we do not directly show the results mentioned in the BG-CMA paper, but the results trained by us. 
    This is because that our code is modified based on BG-CMA, but the computation environment difference cause that we achieve the different results of BG-CMA with claimed.
    Therefore, we choose to show validation results under our unified computation environment, denoted by * here.
    In comparison results, the improvements of our Graph-CMA in SR are 3 $\%$ absolute (7$\%$ relative ) in val-seen and 3 $\%$ absolute (8$\%$ relative) in val-unseen compared with BG-CMA. The improvements of our Graph-CMA in SPL are 4 $\%$ absolute (10.5$\%$ relative ) in val-seen and 3 $\%$ absolute (9.4$\%$ relative) in val-seen compared with BG-CMA. The comparison shows that our method enhances the navigation performance and generalization ability of the agent.
    
    \begin{table}[t]\small
    	\centering
    	\caption{Quantitative Comparison showing the effect of our ERG.}
    	\label{Experiment3}
    	\begin{tabular}{c|cccccccccc}
    		\toprule 
    		&Val-unseen\\
    		Methods&TL$\downarrow$\quad NE$\downarrow$\quad OSR$\uparrow$\quad SR$\uparrow$\quad SPL$\uparrow$\\
    		\midrule
    		*CMA&8.64\quad\ 7.37\quad\ 40\quad\ 32\quad\ 30\quad\ \\
    		*CMA+ERG& \textbf{8.16}\quad\ \textbf{6.98}\quad\ \textbf{43}\quad\ \textbf{35}\quad\ \textbf{33}\quad\ \\

    		\midrule
    		*LAW& \textbf{9.87}\quad\ 7.26\quad\ 58\quad\ 35\quad\ 31\quad\ \\
    		*LAW+ERG&9.95\quad\ \textbf{7.18}\quad\ \textbf{59}\quad\ \textbf{37}\quad\ \textbf{32}\quad\ \\

    		\midrule
    		*BG-CMA& \textbf{9.21}\quad\ 6.55\quad\ 42\quad\ 36\quad\ 32\quad\ \\
    		*BG-CMA+ERG&9.96\quad\ \textbf{6.20}\quad\ \textbf{48}\quad\ \textbf{39}\quad\ \textbf{35}\quad\ \\
    		
    		\bottomrule 
    	\end{tabular} 
    \end{table}

\subsection{Quantitative Comparison with ERG}
    To reveal the effectiveness of ERG, we design an quantitative comparison for the ERG. To be specific, we do comparison between the baselines and baselines with our proposed ERG. Table \ref{Experiment3} shows that our proposed ERG significantly improves the baselines performances. In the comparison about CMA baseline in Val-unseen, CMA+ERG improves 3$\%$ absolute (9.4$\%$ relative) in SR and 3$\%$ absolute (10$\%$ relative) in SPL. Notice CMA method focuses on cross-modal attention spatial visual reasoning ability. The improvement reveals ERG has better cross-modal matching and visual reasoning ability than CMA. In the comparison about LAW baseline in Val-unseen, LAW+ERG improves 2$\%$ absolute (5.7$\%$ relative) in SR and 1$\%$ absolute (3.2$\%$ relative) in SPL. Notice LAW method focuses
    on language understanding ability during the trajectory. The improvement reveals ERG has better language understanding than LAW. In the comparison about BG-CMA baseline in Val-unseen, apart from improvements in SR and SPL, BG-CMA+ERG also improves  6$\%$ absolute (14.3$\%$ relative) in OSR. Notice OSR measures success for each agent under oracle stopping rule, i.e whether agent stops at closest point to the goal on its trajectory. The huge improvement in this metric shows that ERG can stop  more precisely when it is close to destination. This further demonstrates the ERG's strong environment understanding capability. 
    
    \begin{figure}[t]
    	\includegraphics[scale=0.41]{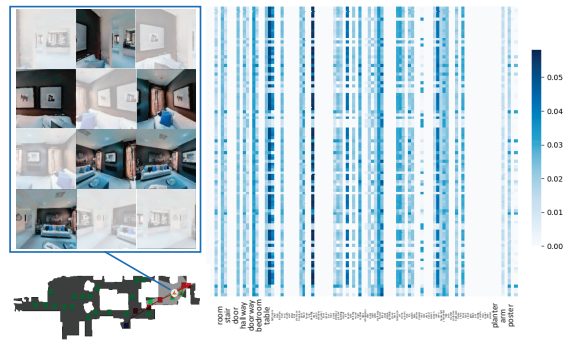}
    	\centering
    	\caption{The agent's understanding of the environment during navigation.} 
    	\label{fig5}
    \end{figure}

\subsection{Qualitative Analysis}
    For a more intuitive view of how our method works for the VLN-CE task, we visualize an qualitative example of agent's trajectory in Figure 5. The bottom left part of the figure is the platform of agent's trajectory. The upper left part of the figure is the candidate waypoint views. 
    The right part of the figure is the expression of ERG generated by GCN as the attention map for interest objects. 
    Objects are represented by column vectors. 
    In the trajectory, the agent successfully reaches the target destination, with a comprehensive understanding of the environment. 
    It first picks 6 waypoint views from images and builds ERG from semantic information. 
    Then from the graph updated, the agent gains the attention of the objects in the environment. 
    From the attention map we can see that ERG shows strong cross-modal matching ability.

\section{Conclusion}
    In this paper, we focus on the agent's ability to understand the environment for VLN-CE and propose a new environment representation.
    First, we introduce semantic information to construct an ERG based on object detection results. 
    Then, GCN is used to learn the relational representation of object-object and object-agent in ERG.
    Finally, the environment representation is obtained by combining the ERG with object label embeddings.
    In order to embed the ERG into the navigation, a novel cross-modal attention navigation framework is proposed with a loss dedicated for ERG. 
    Experiments along with further analysis validate the cross-modal matching and strong generalization ability of our proposed environment representation and the proposed framework.

\bibliographystyle{named}
\bibliography{arxiv}

\end{document}